\documentclass[conference]{IEEEtran}
\IEEEoverridecommandlockouts
\usepackage{cite}
\usepackage{amsmath,amssymb,amsfonts}
\usepackage{algorithmic}
\usepackage{graphicx}
\usepackage{textcomp}
\usepackage{xcolor}
\usepackage{booktabs}
\usepackage{multirow}
\def\BibTeX{{\rm B\kern-.05em{\sc i\kern-.025em b}\kern-.08em
    T\kern-.1667em\lower.7ex\hbox{E}\kern-.125emX}}

\begin{document}

\title{Automated Wicket-Taking Delivery Segmentation and Trajectory-Based Dismissal-Zone Analysis in Cricket Videos Using OCR-Guided YOLOv8}

\author{
\IEEEauthorblockN{
Joy Karmoker\IEEEauthorrefmark{1},
Masum Billah\IEEEauthorrefmark{1},
Mst Jannatun Ferdous\IEEEauthorrefmark{1},
Akif Islam\IEEEauthorrefmark{1},
Mohd Ruhul Ameen\IEEEauthorrefmark{2},
Md. Omar Faruqe\IEEEauthorrefmark{1}
}
\IEEEauthorblockA{\IEEEauthorrefmark{1}Department of Computer Science and Engineering, University of Rajshahi, Rajshahi, Bangladesh}
\IEEEauthorblockA{\IEEEauthorrefmark{2}Marshall University, Huntington, WV, USA}
\IEEEauthorblockA{
\texttt{joykarmoker27@gmail.com},
\texttt{masumbillah3772@gmail.com},
\texttt{ferdousjannatun50@gmail.com},\\
\texttt{iamakifislam@gmail.com},
\texttt{ameen@marshall.edu},
\texttt{faruqe.cse@gmail.com}
}
}

\maketitle

\begin{abstract}
Cricket generates a rich stream of visual and contextual information, yet much of its tactical analysis still depends on slow and subjective manual review. Motivated by the need for a more efficient and data-driven alternative, this paper presents an automated approach for cricket video analysis that identifies wicket-taking deliveries, detects the pitch and ball, and models ball trajectories for post-match assessment. The proposed system combines optical character recognition (OCR) with image preprocessing techniques, including grayscale conversion, power transformation, and morphological operations, to robustly extract scorecard information and detect wicket events from broadcast videos. For visual understanding, YOLOv8 is employed for both pitch and ball detection. The pitch detection model achieved 99.5\% mAP50 with a precision of 0.999, while the transfer learning-based ball detection model attained 99.18\% mAP50 with 0.968 precision and 0.978 recall. Based on these detections, the system further models ball trajectories to reveal regions associated with wicket-taking deliveries, offering analytical cues for trajectory-based dismissal-zone interpretation and potential batting vulnerability assessment. Experimental results on multiple cricket match videos demonstrate the effectiveness of the proposed approach and highlight its potential for supporting coaching, tactical evaluation, and data-driven decision-making in cricket.
\end{abstract}

\begin{IEEEkeywords}
Cricket video analysis, YOLOv8, object detection, trajectory modeling, OCR, deep learning, sports analytics.
\end{IEEEkeywords}
\section{Introduction}

Automatic sports video analysis supports indexing, summarization, tracking, and tactical assessment from large volumes of broadcast footage \cite{naik2022review}. In cricket, match videos contain valuable cues about wicket patterns and delivery behavior, yet extracting such information still depends heavily on manual review. This process is time-consuming, subjective, and difficult to scale across matches and players.

Broadcast cricket video remains challenging for automated analysis because it contains camera motion, zoom variation, motion blur, occlusion, and large object-scale changes \cite{shih2017survey}. These challenges are especially severe for cricket-ball localization, since the ball is small and fast-moving. Moreover, practical analysis requires not only object detection, but also identification of wicket-taking segments and interpretation of their associated trajectories.

A useful cue in broadcast sports video is the superimposed scoreboard or scorecard, which provides compact semantic information about match progression \cite{guo2011scoreboard}. In cricket, score changes and wicket updates can help identify important moments from long videos \cite{shukla2018highlight}. This motivates a practical strategy in which scorecard information is first used to locate wicket events, after which detailed visual analysis is applied only to the corresponding delivery segments.

Recent object detection models have made high-precision sports video analysis increasingly feasible. YOLOv8 is particularly suitable for this task because it provides a strong balance between accuracy and efficiency and can handle both large structured regions such as the pitch and small fast-moving objects such as the ball \cite{reis2024yolov8}. This makes it a practical choice for combining OCR-guided event localization with broadcast-video object detection.

Motivated by these challenges, this paper presents an integrated framework for wicket-focused cricket video analysis. The proposed pipeline first applies OCR to the broadcast scorecard region to detect wicket transitions and extract relevant delivery segments. YOLOv8-based detectors are then used to localize the pitch and cricket ball, after which ball positions are used to model delivery trajectories and visualize dismissal-associated regions for post-match analysis.

The main contributions of this paper are summarized as follows:
\begin{enumerate}
    \item We propose an OCR-guided wicket segmentation framework that analyzes broadcast scorecards to automatically identify wicket events and extract wicket-taking delivery segments from cricket videos.
    
    \item We develop YOLOv8-based detection models for pitch and cricket-ball localization, enabling accurate spatial understanding of the delivery region in challenging broadcast footage.
    
    \item We introduce a trajectory modeling and visualization stage that aggregates wicket-associated ball paths to highlight dismissal-zone patterns and potential batting vulnerability cues.
    
    \item We present an end-to-end cricket analytics pipeline that connects semantic event identification, object detection, and trajectory-based interpretation within a single automated framework for post-match analysis.
\end{enumerate}

\section{Related Work}

\subsection{Sports Video Analysis and Cricket Event Understanding}
Automatic sports video understanding has been widely studied for tasks such as event detection, highlight generation, semantic indexing, and object tracking from broadcast videos \cite{shih2018survey,naik2022review}. In cricket, earlier work focused mainly on summarization and event-centric indexing rather than delivery-level analysis. Narasimhan \textit{et al.} proposed a genetic-algorithm-based method for cricket video summarization, while Kolekar and Sengupta explored semantic concept mining for automatic highlight generation from cricket broadcasts \cite{narasimhan2010genetic,kolekar2010semantic}. Goyani \textit{et al.} further investigated hierarchical semantic event detection for cricket video indexing \cite{goyani2011keyframe}, and Shukla \textit{et al.} later combined event-driven and excitement-based features for automatic cricket highlight generation \cite{shukla2018highlight}. While these studies demonstrate the value of automated cricket video analysis, they mainly address summarization and highlight extraction rather than wicket-focused delivery segmentation for tactical interpretation.

\subsection{Scoreboard-Guided Video Understanding}
Another relevant research direction has explored the use of superimposed scoreboards or scorecards in sports broadcasts. Because score overlays provide compact and continuously updated contextual information, they serve as useful cues for video indexing and alignment. Guo \textit{et al.} proposed a scoreboard localization and recognition framework based on SIFT matching and OCR enhancement, showing that on-screen score information can act as a stable semantic anchor in sports videos \cite{guo2011scoreboard}. More recently, Agarwal \textit{et al.} introduced the ASAP framework, which parses embedded scorecards using OCR and aligns sports videos with external annotations; in cricket, the over and ball indicator was used as a temporal reference for semantic alignment \cite{agarwal2023asap}. Although these studies confirm the usefulness of scorecard information for indexing and annotation, they do not directly address wicket-taking delivery extraction or its integration with later spatial analysis.

\subsection{Ball Tracking and Object Localization in Sports}
Ball tracking and localization have also been widely studied in sports vision, although they remain difficult because the ball is often small, fast-moving, blurred, and occluded. Yan \textit{et al.} proposed a tennis ball tracking method for low-quality single-camera broadcast videos \cite{yan2005tennis}, while Wang \textit{et al.} showed that contextual reasoning about team play can improve ball-tracking performance in challenging scenarios \cite{wang2014balltracking}. In cricket, Arora \textit{et al.} developed a low-cost umpire-assistance and ball-tracking system using a single smartphone camera, but the method was designed for constrained capture conditions rather than unconstrained broadcast footage \cite{arora2017smartphone}. Overall, these studies highlight the importance of accurate object localization in sports analytics, but they do not provide an integrated solution for wicket-event extraction, pitch localization, and delivery trajectory analysis from broadcast cricket videos.

\subsection{Research Gap}
Despite progress in cricket video summarization, scoreboard-aware video understanding, and sports ball tracking, an important gap remains at the intersection of these areas. Existing studies generally treat event detection, score parsing, and object tracking as separate problems, whereas practical cricket analytics requires them to work together in a unified pipeline. In particular, limited attention has been given to automatically extracting wicket-taking delivery segments from broadcast videos and linking them with pitch-ball localization and trajectory-based interpretation. This limitation is important for post-match tactical review, where analysts need both event-level understanding and spatial analysis of the corresponding delivery. The present work addresses this gap by integrating OCR-guided wicket-event segmentation with YOLOv8-based pitch and ball detection, followed by trajectory-based dismissal-zone analysis within a single automated framework.

\section{Methodology}

\begin{figure}[!b]
\centering
\includegraphics[width=\columnwidth]{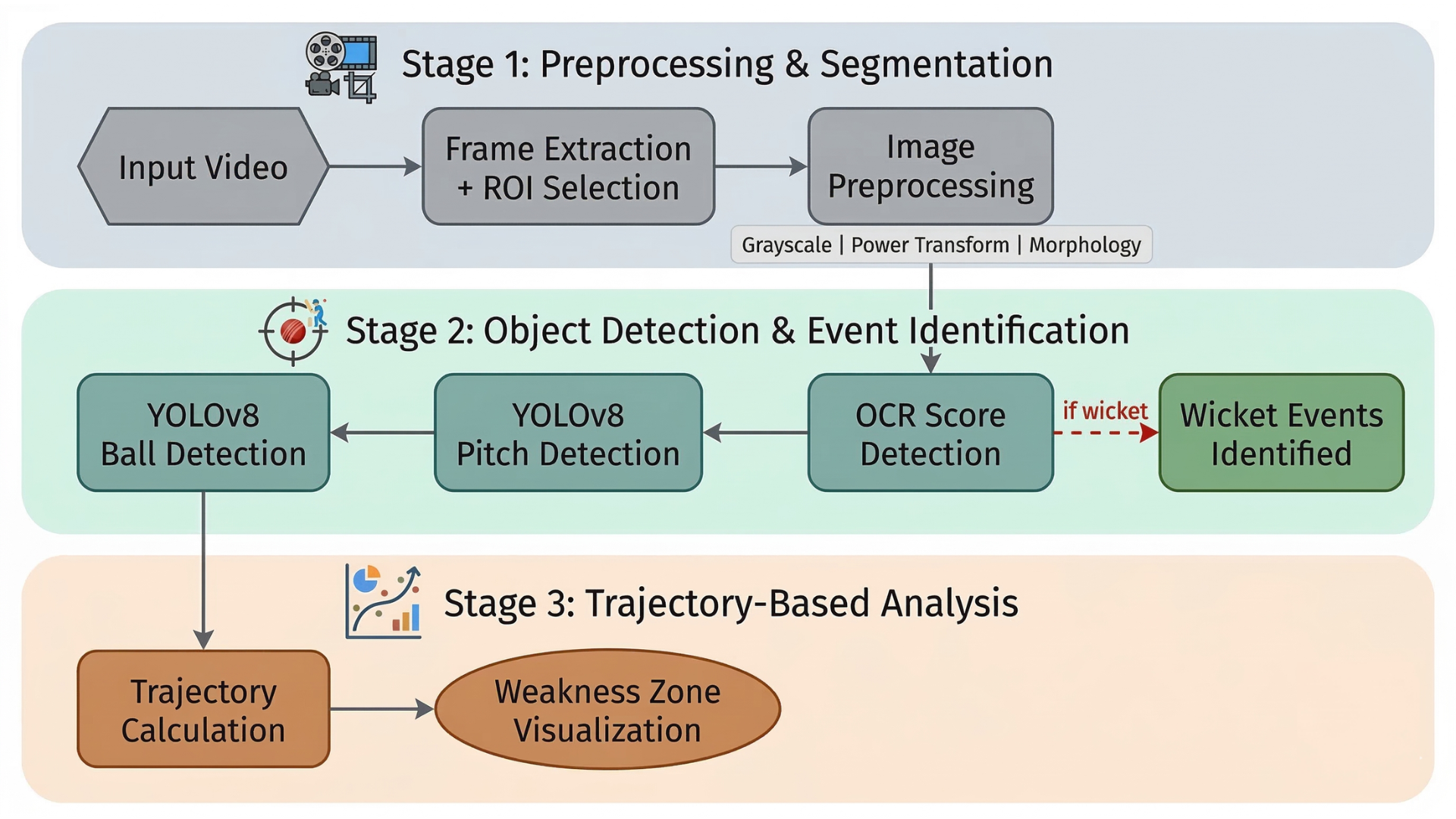}
\caption{Overview of the proposed framework. The pipeline consists of three sequential stages: OCR-guided wicket-event segmentation, YOLOv8-based pitch and ball detection, and trajectory-based dismissal-zone analysis.}
\label{fig:methodology}
\end{figure}

\subsection{Overall Framework}
The proposed methodology is organized into three consecutive stages, as illustrated in Fig.~\ref{fig:methodology}. First, wicket-taking delivery segments are identified from broadcast videos through scorecard monitoring and OCR-based event detection. Second, the extracted segments are processed using YOLOv8 models to localize the pitch and the cricket ball. Finally, the detected ball positions are used to model delivery trajectories and visualize wicket-associated spatial patterns. This staged design allows semantic event localization, spatial detection, and post-event interpretation to be integrated within a unified analysis pipeline.

\begin{figure*}[!t]
\centering
\includegraphics[width=0.9\textwidth]{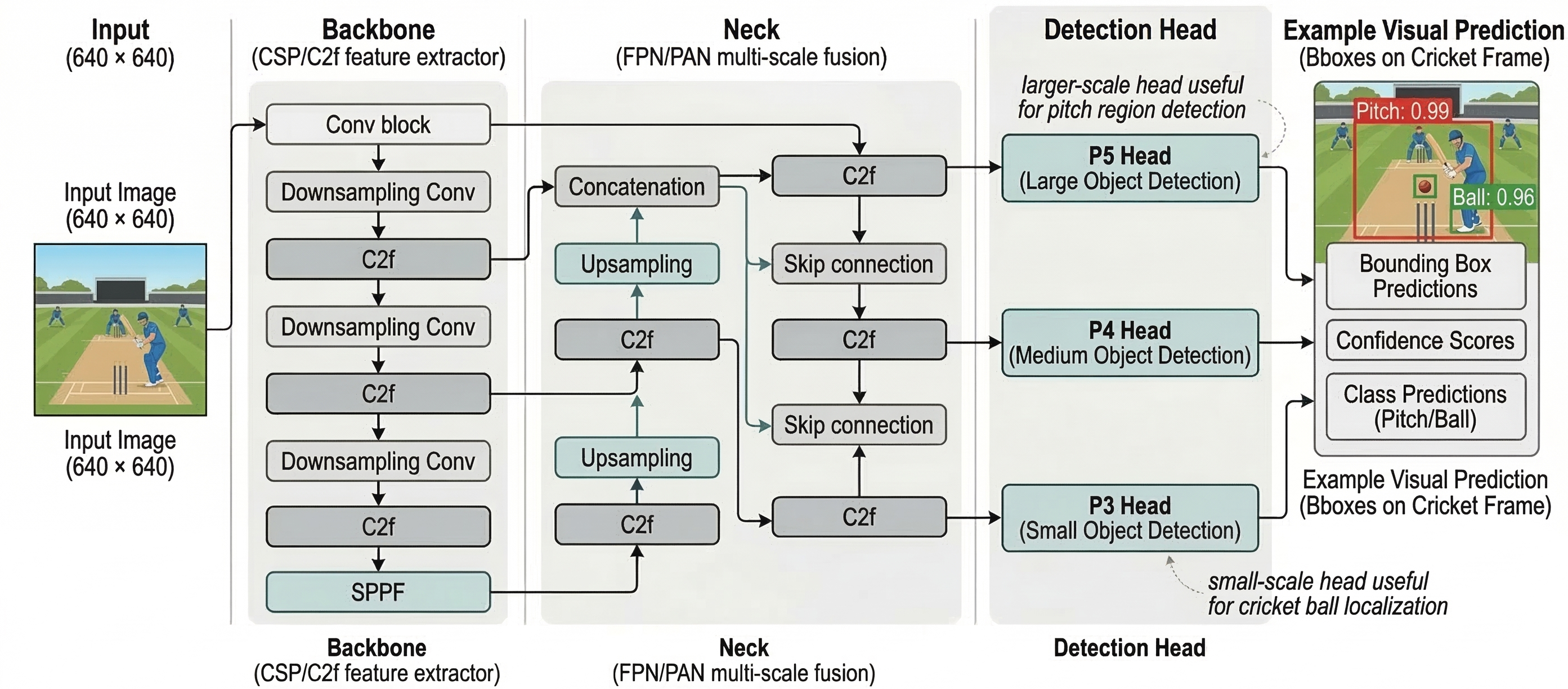}
\caption{High-level YOLOv8 architecture adopted for pitch and ball detection. The model consists of a backbone for hierarchical feature extraction, a neck for multi-scale feature aggregation, and detection heads for object localization and confidence prediction.}
\label{fig:yolov8_arch}
\end{figure*}

\subsection{OCR-Guided Wicket Event Segmentation}

\subsubsection{Frame Sampling and Scorecard Localization}
The wicket segmentation stage begins by sampling frames from the input broadcast video at fixed temporal intervals, with a default interval of 0.1~s. For each sampled frame, a predefined region of interest (ROI) is used to isolate the scorecard area, which is typically located near the lower portion of the frame. The scorecard provides the current match score in compact textual form, where runs and wickets are separated by either a hyphen or a forward slash. This score representation serves as the primary signal for identifying wicket transitions.

\subsubsection{Preprocessing for OCR Enhancement}
To improve text readability prior to OCR, a sequence of image preprocessing operations is applied to the scorecard ROI. First, the image is converted from BGR to grayscale using
\begin{equation}
\text{Gray} = 0.299R + 0.587G + 0.114B.
\end{equation}
Next, a power transformation is performed to enhance contrast:
\begin{equation}
I_{\text{output}} = 255 \times \left(\frac{I_{\text{input}}}{255}\right)^\gamma,
\end{equation}
where $\gamma = 7$ in the present implementation, followed by intensity inversion. Finally, morphological operations, including dilation and erosion with a $15 \times 15$ kernel, are applied to suppress noise while preserving digit structure. A median blur is then used to smooth residual artifacts. This preprocessing sequence improves OCR robustness under typical broadcast conditions such as compression noise, overlays, and illumination variation.

\subsubsection{Score Parsing and Wicket Boundary Detection}
The preprocessed scorecard text is passed to an OCR engine and subsequently validated using regular-expression rules corresponding to common score formats, namely \texttt{Wicket-Runs}, \texttt{Wicket/Runs}, \texttt{Runs-Wickets}, and \texttt{Runs/Wickets}. A wicket event is declared when the recognized wicket count increases relative to the preceding valid score instance. Once such an increment is detected, the corresponding delivery segment is extracted and saved for downstream spatial analysis. In this way, OCR serves not merely as a text-reading component, but as a temporal event trigger for wicket-focused video segmentation.

\subsection{Pitch and Ball Localization with YOLOv8}

\subsubsection{Detection Architecture}
Pitch and ball detection are performed using YOLOv8 due to its favorable trade-off between localization accuracy and computational efficiency. As shown in Fig.~\ref{fig:yolov8_arch}, the architecture consists of three principal components: a backbone, a neck, and detection heads. The backbone extracts hierarchical visual representations from the input frame, the neck aggregates multi-scale information to improve representation quality across object sizes, and the detection heads predict bounding boxes and associated confidence scores. This design is particularly suitable for the present task because it must jointly handle a relatively large structured region, namely the pitch, and a small fast-moving object, namely the cricket ball.

\subsubsection{Detection Datasets}
Two separate datasets were used for detector development. The pitch detection dataset contains 951 annotated images obtained from Roboflow \cite{roboflow_pitch_2023}, while the ball detection dataset contains 257 annotated images from Roboflow \cite{roboflow_ball_2023}. In both cases, annotations are provided in YOLO format with a single target class per dataset. The pitch dataset emphasizes stable localization of a large scene component, whereas the ball dataset focuses on small-object detection under motion blur, scale variation, and diverse broadcast conditions. The main dataset characteristics are summarized in Table~\ref{tab:dataset_comprehensive}.

\begin{table}[!t]
\centering
\caption{Summary of the pitch and ball detection datasets used in the proposed framework.}
\begin{tabular}{@{}lcc@{}}
\toprule
\textbf{Characteristic} & \textbf{Pitch Detection} & \textbf{Ball Detection} \\
\midrule
\multicolumn{3}{l}{\textit{Dataset Split}} \\
\quad Total Images & 951 & 257 \\
\quad Training Set & 797 (83.8\%) & 210 (81.7\%) \\
\quad Validation Set & 130 (13.7\%) & 28 (10.9\%) \\
\quad Test Set & 12 (1.3\%) & 27 (10.5\%) \\
\midrule
\multicolumn{3}{l}{\textit{Image Properties}} \\
\quad Resolution & 640 × 640 px & 640 × 640 px \\
\quad Format & JPG & JPG \\
\quad Color Space & RGB & RGB \\
\quad Avg. File Size & 85 KB & 92 KB \\
\midrule
\multicolumn{3}{l}{\textit{Annotation Details}} \\
\quad Annotation Format & YOLOv8 & YOLOv8 \\
\quad Number of Classes & 1 (Pitch) & 1 (Ball) \\
\quad Avg. Objects/Image & 1.0 & 1.2 \\
\quad Bounding Box Type & Rectangle & Rectangle \\
\quad Labeling Tool & Roboflow & Roboflow \\
\midrule
\multicolumn{3}{l}{\textit{Data Characteristics}} \\
\quad Object Size (avg.) & Large (60-80\%) & Small (1-3\%) \\
\quad Aspect Ratio Var. & Low & Medium \\
\quad Lighting Conditions & Varied & Varied \\
\quad Camera Angles & Multiple & Multiple \\
\midrule
\multicolumn{3}{l}{\textit{Augmentation Applied}} \\
\quad Horizontal Flip & Yes & Yes \\
\quad Rotation & ±15° & ±10° \\
\quad Brightness Adj. & ±20\% & ±15\% \\
\quad Noise Addition & Gaussian & Gaussian \\
\bottomrule
\end{tabular}
\label{tab:dataset_comprehensive}
\end{table}

\subsubsection{Model Training Strategy}

The pitch detector was optimized through three successive training configurations to improve convergence stability and generalization. These configurations varied in optimizer choice, batch size, number of frozen layers, and loss weighting. For ball detection, both training from scratch and transfer learning with COCO-pretrained weights were investigated. The transfer-learning setting produced substantially stronger performance, indicating that pretrained visual representations are especially beneficial for small-object localization in broadcast cricket footage. Overall, the training strategy was designed to balance detector robustness with the distinct visual characteristics of the two target classes.

\subsection{Trajectory-Based Dismissal-Zone Analysis}
After detection, ball centers are collected from frames in which the detected ball lies within the detected pitch region. These spatial observations are then arranged in temporal order to approximate the delivery trajectory over the pitch plane. By overlaying trajectories from multiple wicket-taking deliveries, the framework produces an aggregated spatial representation of dismissal-associated regions. This trajectory visualization is intended to provide interpretable cues regarding recurring dismissal-zone patterns and possible batting vulnerability trends during post-match review.

\section{Results and Discussion}

\begin{table}[!t]
\centering
\caption{Model Performance Summary and Comparison}
\begin{tabular}{@{}lcccc@{}}
\toprule
\textbf{Model} & \textbf{Precision} & \textbf{Recall} & \textbf{mAP50} & \textbf{mAP50-95} \\
\midrule
\multicolumn{5}{l}{\textit{Pitch Detection Models}} \\
\quad Training 1 (Adam) & 0.999 & 1.000 & 0.995 & 0.984 \\
\quad Training 2 (SGD, 18F) & 0.999 & 1.000 & 0.995 & 0.982 \\
\quad Training 3 (SGD, 22F) & 0.959 & 0.908 & 0.969 & 0.709 \\
\midrule
\multicolumn{5}{l}{\textit{Ball Detection Models}} \\
\quad From Scratch & 0.725 & 0.725 & 0.773 & 0.374 \\
\quad Transfer Learning & \textbf{0.968} & \textbf{0.978} & \textbf{0.9918} & \textbf{0.930} \\
\midrule
\multicolumn{5}{l}{\textit{Improvement Analysis}} \\
\quad TL vs. Scratch (\%) & +33.5\% & +34.9\% & +28.3\% & +148.7\% \\
\bottomrule
\end{tabular}
\label{tab:performance_summary}
\end{table}

\subsection{Detection Performance Analysis}

Table~\ref{tab:performance_summary} summarizes the overall performance of the pitch and ball detection models. For pitch localization, the first two training configurations produced nearly identical results, both achieving 0.995 mAP50 with almost perfect precision and recall. These findings indicate that the pitch region can be learned reliably from the available dataset and that stable localization is possible under diverse broadcast conditions. By contrast, the third pitch configuration yielded lower performance, particularly in mAP50--95, suggesting that the more aggressive freezing and batch configuration reduced fine-grained localization quality even though detection remained strong at the coarse IoU level.

\subsubsection{Pitch Detection Results}
The detailed pitch detection results are reported in Table~\ref{tab:pitch_results}. Training~1 and Training~2 both achieved near-saturated performance, while Training~3 showed a noticeable decline in recall and mAP50--95. This pattern suggests that, for pitch detection, additional optimization beyond a stable high-performing configuration does not necessarily improve final accuracy. Instead, the results indicate that the task is comparatively well-structured, and that over-adjustment of training parameters may reduce localization precision without providing a practical advantage.

\begin{table}[!b]
\caption{Pitch Detection Results - Final Epochs}
\centering
\small
\begin{tabular}{@{}lcccc@{}}
\toprule
Training & Precision & Recall & mAP50 & mAP50-95 \\
\midrule
Train 1 & 0.999 & 1.000 & 0.995 & 0.984 \\
Train 2 & 0.999 & 1.000 & 0.995 & 0.982 \\
Train 3 & 0.959 & 0.908 & 0.969 & 0.709 \\
\bottomrule
\end{tabular}
\label{tab:pitch_results}
\end{table}

The pitch detector exhibited stable convergence during training, with both optimization losses and evaluation metrics improving rapidly in the early epochs before reaching a consistent plateau. This behavior is aligned with the strong numerical performance reported in Table~\ref{tab:pitch_results} and indicates that the detector learns the pitch region reliably from the available annotations. From an application perspective, such robustness is important because the pitch serves as the reference spatial region for subsequent ball filtering and trajectory construction.


\subsubsection{Ball Detection Results}
Ball detection is substantially more challenging than pitch detection because the cricket ball is small, fast-moving, and often affected by motion blur, background clutter, and scale variation. The comparative results in Table~\ref{tab:ball_results} show a clear advantage for transfer learning over training from scratch. The transfer-learning model achieved 0.968 precision, 0.978 recall, and 0.9918 mAP50, whereas the model trained from scratch reached only 0.773 mAP50. This large gap confirms that pretrained visual representations are highly beneficial for small-object localization in broadcast cricket footage.

\begin{table}[t]
\caption{Ball Detection Results Comparison}
\centering
\small
\begin{tabular}{@{}lcccc@{}}
\toprule
Method & Precision & Recall & mAP50 & mAP50-95 \\
\midrule
From Scratch & 0.725 & 0.725 & 0.773 & 0.374 \\
Transfer Learning & 0.968 & 0.978 & 0.9918 & 0.930 \\
\bottomrule
\end{tabular}
\label{tab:ball_results}
\end{table}

This result is particularly meaningful for the proposed pipeline. Since trajectory estimation depends directly on the consistency of ball localization, improvements in ball detection quality propagate to the final analytical stage. In other words, transfer learning does not merely improve an intermediate metric; it strengthens the reliability of downstream trajectory-based interpretation.

\subsection{Qualitative Assessment of Spatial Localization}

\begin{figure*}[t]
\centering
\includegraphics[width=0.7\textwidth]{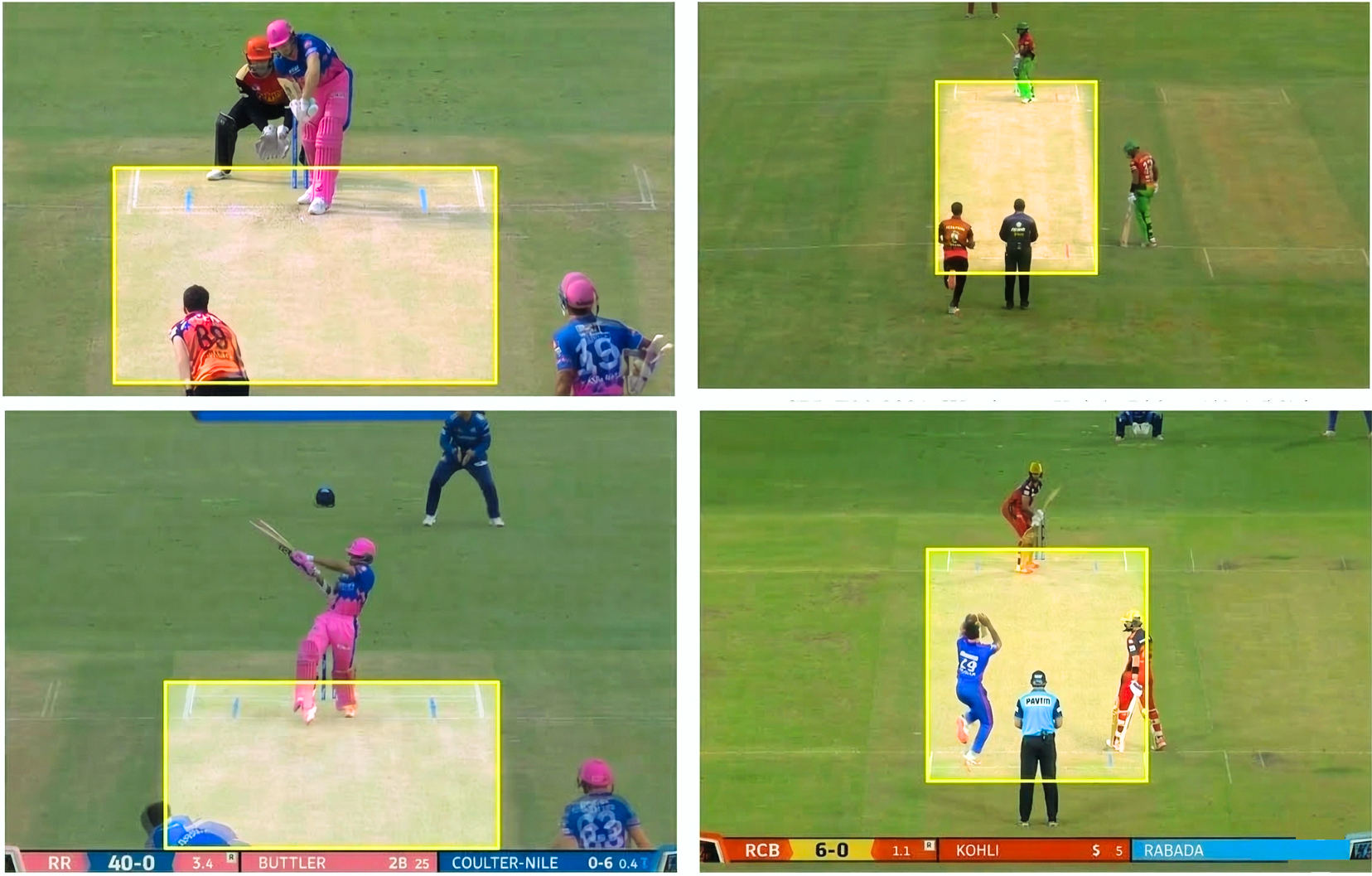}
\caption{Representative qualitative results of pitch detection on broadcast cricket frames. The predicted bounding boxes accurately localize the pitch region across different viewpoints, player configurations, and match scenes.}
\label{fig:detection_results}
\end{figure*}

In addition to the quantitative metrics, Fig.~\ref{fig:detection_results} provides representative qualitative examples of the detection outputs. The predicted bounding boxes closely align with the visible pitch area across different camera viewpoints and batting scenes, demonstrating that the detector generalizes beyond a single broadcast layout. Such qualitative evidence complements the numerical results by showing that the detector remains effective in realistic match frames rather than only in isolated evaluation statistics.

\begin{table}[!t]
\centering
\caption{Training Hyperparameters Across All Experiments}
\scriptsize
\begin{tabular}{@{}lccccc@{}}
\toprule
\textbf{Parameter} & \textbf{Pitch T1} & \textbf{Pitch T2} & \textbf{Pitch T3} & \textbf{Ball Scratch} & \textbf{Ball TL} \\
\midrule
\multicolumn{6}{l}{\textit{Basic Training Settings}} \\
\quad Total Epochs & 500 & 700 & 700 & 700 & 700 \\
\quad Batch Size & 8 & 16 & 30 & 8 & 8 \\
\quad Image Size & 640 & 640 & 640 & 640 & 640 \\
\quad Optimizer & Adam & SGD & SGD & SGD & SGD \\
\quad Patience & 20 & 30 & 30 & 30 & 30 \\
\midrule
\multicolumn{6}{l}{\textit{Learning Rate Configuration}} \\
\quad Initial LR (lr0) & 0.001 & 0.0001 & 0.0001 & 0.01 & 0.001 \\
\quad Final LR (lrf) & 0.01 & 0.01 & 0.01 & 0.01 & 0.01 \\
\quad Momentum & 0.9 & 0.9 & 0.9 & 0.9 & 0.9 \\
\quad Warmup Epochs & 3 & 3 & 3 & 3 & 3 \\
\quad Warmup Bias LR & 0.001 & 0.001 & 0.001 & 0.001 & 0.001 \\
\midrule
\multicolumn{6}{l}{\textit{Transfer Learning Settings}} \\
\quad Pretrained Weights & No & No & No & No & COCO \\
\quad Frozen Layers & 0 & 18 & 22 & 0 & 18 \\
\quad Single Class Mode & Yes & Yes & Yes & Yes & Yes \\
\midrule
\multicolumn{6}{l}{\textit{Loss Function Weights}} \\
\quad Box Loss Weight & 7.5 & 8.5 & 9.0 & 7.5 & 5.5 \\
\quad Class Loss Weight & 0.5 & 0.5 & 0.25 & 0.5 & 0.4 \\
\quad DFL Loss Weight & 1.5 & 0.5 & 0.25 & 1.5 & 0.5 \\
\midrule
\multicolumn{6}{l}{\textit{Final Performance Metrics}} \\
\quad Best Epoch & 116 & 179 & 161 & 173 & 544 \\
\quad Precision & 0.999 & 0.999 & 0.959 & 0.725 & \textbf{0.968} \\
\quad Recall & 1.000 & 1.000 & 0.908 & 0.725 & \textbf{0.978} \\
\quad mAP50 & 0.995 & 0.995 & 0.969 & 0.773 & \textbf{0.9918} \\
\quad mAP50-95 & 0.984 & 0.982 & 0.709 & 0.374 & \textbf{0.930} \\
\midrule
\multicolumn{6}{l}{\textit{Validation Loss (Final Epoch)}} \\
\quad Box Loss & 0.279 & 0.255 & 0.281 & 1.261 & 0.255 \\
\quad Class Loss & 0.189 & 0.116 & 0.476 & 0.749 & 0.163 \\
\quad DFL Loss & 0.875 & 0.269 & 0.298 & 1.014 & 0.267 \\
\bottomrule
\end{tabular}
\label{tab:training_comprehensive}
\end{table}

The complete training settings and corresponding outcomes are summarized in Table~\ref{tab:training_comprehensive}. A comparison across experiments presents that the proposed framework is sensitive to training strategy, especially for ball detection. While pitch detection remains robust across multiple configurations, ball detection benefits substantially from pretrained initialization and a more carefully controlled optimization setup. This difference reflects the relative difficulty of the two localization tasks and supports the design choice of treating them as separate detection problems rather than forcing a single unified detector.

\subsection{Wicket Event Segmentation Performance}

The OCR-guided scorecard analysis was effective in identifying wicket-taking delivery segments across multiple match videos and scorecard layouts. The preprocessing pipeline, particularly the combination of grayscale conversion, power transformation, and morphological enhancement, improved text readability and reduced the impact of low contrast and overlay artifacts. As a result, score transitions corresponding to wicket increments could be detected with sufficient reliability to isolate semantically relevant delivery segments for the downstream detection stage.

From a system perspective, this segmentation stage is important because it narrows the analysis to event-relevant video portions rather than processing the entire match uniformly. This reduces unnecessary computation and aligns the later spatial analysis with tactically meaningful events. At the same time, the dependence on visible scorecards implies that segmentation quality may vary when the broadcast overlay is partially occluded, reformatted, or temporarily absent. Thus, while the OCR-driven strategy is practical and effective for offline broadcast analysis, it also defines one of the main operational constraints of the proposed framework.

\subsection{System-Level Interpretation}

Taken together, the results indicate that the proposed framework is effective as an end-to-end offline cricket video analysis pipeline. The OCR module provides event-focused temporal filtering, the YOLOv8 detectors deliver accurate spatial localization, and the resulting detections enable trajectory-based post-match interpretation. The strongest evidence in this study comes from the detection stages, where both pitch and ball localization achieved high accuracy, particularly under the transfer-learning setting.

However, the system should be interpreted as a practical analytical framework rather than a fully comprehensive cricket intelligence solution. The final trajectory-based stage is intended to reveal wicket-associated spatial tendencies that may support tactical review, not to establish definitive causal claims about batting weakness in isolation. Match context, bowler strategy, pitch condition, and batter intent are not explicitly modeled in the present work. Nevertheless, the results demonstrate that semantically guided video segmentation combined with accurate object detection can provide a useful foundation for interpretable cricket analytics in a broadcast-video setting.

\section{Conclusion}
This paper presented an automated framework for cricket video analysis that integrates OCR-guided wicket-event segmentation, YOLOv8-based spatial localization, and trajectory-based post-match interpretation within a unified pipeline. The proposed approach is intended for offline broadcast-video analysis and remains dependent on visible scorecards, consistent broadcast overlays, and stable visual conditions, which may limit performance under occlusion, layout variation, or missing contextual cues. Nevertheless, the framework demonstrates the practical value of combining semantic event identification with object detection for cricket analytics. Future work will focus on quantitative evaluation of wicket-event segmentation, improved dismissal-zone visualization, incorporation of richer match context, and optimization toward near-real-time deployment.

\bibliographystyle{IEEEtran}
\bibliography{references}

\end{document}